\documentclass[letterpaper, 10 pt, conference]{ieeeconf}  
\usepackage{graphicx}
\IEEEoverridecommandlockouts                              

\usepackage{amsmath}
\usepackage{amssymb}
\usepackage{graphics}
\usepackage[hidelinks]{hyperref}
\usepackage{xcolor}
\usepackage{dblfloatfix}
\usepackage {tikz}
\usepackage{mathtools}
\usepackage{svg}
\usepackage[bottom]{footmisc}
\usepackage{amsmath}
\usepackage{soul}
\usepackage[switch]{lineno}
\usepackage{multirow}
\usepackage{caption}
\usepackage[noend]{algpseudocode}
\usepackage[linesnumbered,ruled,vlined, noend]{algorithm2e}
\usetikzlibrary {positioning}

\title{\LARGE \bf
Collaborative Adaptation: Learning to Recover from Unforeseen Malfunctions in Multi-Robot Teams
}

\author{Yasin Findik$^{1}$, Paul Robinette$^{2}$, Kshitij Jerath$^{3}$, and  S. Reza Ahmadzadeh$^{1}$
\thanks{$^{1}$ PeARL Lab, Richard Miner School of Computer and Information Sciences, University of Massachusetts Lowell, MA, USA {\tt\small yasin\_findik@student.uml.edu, reza@cs.uml.edu}}%
\thanks{$^{2}$ Department of Electrical and Computer Engineering, University of Massachusetts Lowell, MA, USA {\tt\small paul\_robinette@uml.edu}}%
\thanks{$^{3}$ Department of Mechanical Engineering, University of Massachusetts Lowell, MA, USA {\tt\small kshitij\_jerath@uml.edu}}
}%
\begin{document}

\maketitle
\thispagestyle{empty}
\pagestyle{empty}

\begin{abstract}

Cooperative multi-agent reinforcement learning (MARL) approaches tackle the challenge of finding effective multi-agent cooperation strategies for accomplishing individual or shared objectives in multi-agent teams. In real-world scenarios, however, agents may encounter unforeseen failures due to constraints like battery depletion or mechanical issues. Existing state-of-the-art methods in MARL often recover slowly -- if at all -- from such malfunctions once agents have already converged on a cooperation strategy. To address this gap, we present the Collaborative Adaptation (CA) framework. CA introduces a mechanism that guides collaboration and accelerates adaptation from unforeseen failures by leveraging inter-agent relationships. Our findings demonstrate that CA enables agents to act on the knowledge of inter-agent relations, recovering from unforeseen agent failures and selecting appropriate cooperative strategies.

\end{abstract}

\section{Introduction}

Multi-robot\footnote[1]{The terms 'robot' and 'agent' are used interchangeably throughout this paper.} scenarios are commonly encountered in various domains, including search \& rescue operations~\cite{kleiner2006rfid}, autonomous driving~\cite{pendleton2017perception, Kim2022CACC}, and logistics \& transportation~\cite{diaz2021editorial}. The coordination and cooperation between agents are essential in these scenarios, enabling them to achieve shared or individual goals~\cite{busoniu2008comprehensive}. They become particularly crucial when addressing unexpected malfunctions that robots may experience, such as battery failure leading to immobilization or rotation failure restricting movement to a single direction. It is imperative for agents to effectively cooperate with each other, autonomously recover from such failures promptly, and adapt their strategies as a team to overcome the challenges arising from agent malfunction(s).

Within the field of Multi-Agent Reinforcement Learning (MARL) for cooperative tasks, the Centralized Training with Decentralized Execution (CTDE) paradigm has emerged as a prominent approach. It effectively addresses a range of cooperative challenges, including curse of dimensionality~\cite{shoham2007if, Haeri2020Swarm}, non-stationarity~\cite{busoniu2008comprehensive}, and global exploration~\cite{matignon2012independent}. Despite its impressive performance in coordination tasks, CTDE-based approaches suffer from a notable drawback: slow adaptation to unexpected agent failures. This issue arises from two primary factors. Firstly, these approaches lack explicit mechanisms to handle such unpredictable failure cases. Secondly, they do not incorporate features that promote enhanced collaboration between agents, resulting in a slower adaptation process where the model must independently discover which collaboration strategies to pursue after learning new ones.

In this paper, we introduce a novel algorithm that extends the CTDE paradigm. Our algorithm leverages a relational network~\cite{findik2023impact} to capture the relative importance assigned by agents to one another, enabling faster adaptation in the face of unexpected robot failures. To evaluate the effectiveness of our method, we  experimented in a multi-robot environment, focusing on a cooperative task with simulated random malfunctions. We compared our approach to  the state-of-the-art, Value Decomposition Networks (VDN)~\cite{sunehag2017value}. The findings of our study demonstrate that our proposed approach facilitates effective cooperation within a multi-robot team, enabling faster adaptation to unforeseen malfunctions through the utilization of relational networks.


\section{Related Work}

In recent years, Multi-Agent Reinforcement Learning (MARL) has emerged as a prominent research area, particularly in cooperative settings. Numerous approaches have been explored to enable effective collaboration among agents in pursuit of a common objective. One widely studied approach is fully centralized learning, where a single controller is shared among all agents, allowing them to learn a joint policy or value function collectively~\cite{claus1998dynamics}. Despite its potential advantages, fully centralized learning can be computationally demanding and face intractability challenges due to the exponential growth of the observation and action space as the number of agents increases. An alternative strategy in MARL is fully decentralized learning, where each agent independently learns its own policy. The cooperative behavior then emerges from the application of these learned policies within the environment. For example, Independent Q-Learning (IQL)~\cite{tan1993multi} employs separate action-value tables for each agent, utilizing Q-learning as the underlying learning mechanism. To address the limitations of tabular Q-learning in high-dimensional state and action spaces, the IQL framework was later extended to incorporate function approximation techniques~\cite{tampuu2017multiagent}. However, independent learning approaches in multi-agent settings are prone to non-stationarity issues, which arise from the changing actions of other agents as perceived by a given agent. Due to the violation of the Markov property in non-stationary environments, the convergence of decentralized algorithms based on Q-learning cannot be guaranteed~\cite{hernandez2017survey}. 

In cooperative MARL scenarios, the limitations of fully centralized and fully decentralized learning approaches have led to the development of a novel paradigm known as Centralized Training with Decentralized Execution (CTDE)~\cite{oliehoek2008optimal}. CTDE enables individual agents to execute their actions autonomously while leveraging a centralized mechanism to integrate their strategies, thereby facilitating effective coordination and alignment towards a common objective. By employing centralized training, CTDE effectively addresses the challenge of non-stationarity in decentralized learning, while also overcoming the scalability challenges associated with centralized learning through decentralized execution. This paradigm has been implemented using two main approaches: policy-based and value-based methods. Policy-based methods such as Multi-Agent Deep Deterministic Policy Gradient (MADDPG)~\cite{lowe2017multi} and Multi-Agent Proximal Policy Optimization (MAPPO)~\cite{yu2021surprising} incorporate a critic that takes into account the global observations of all agents. On the other hand, value-based techniques including Value Decomposition Networks ~\cite{sunehag2017value}, QMIX~\cite{rashid2020monotonic}, and QTRAN~\cite{son2019qtran} enhance Q-Learning by incorporating a centralized function that calculates the joint Q-value based on the individual action-values of each agent. These approaches have demonstrated effectiveness in addressing challenges related to multi-agent coordination and have shown superior performance across a range of scenarios.

Existing research in cooperative MARL has primarily focused on achieving optimal solutions, ranging from fully centralized learning to the CTDE paradigm. However, when unforeseen failures occur during the execution of learned behaviors, these approaches may not promptly adapt the agents' policies. One possible approach to recover from robot malfunctions is to predict the malfunctioning robot and its timing by enabling agents to estimate the actions of other agents. The concept of LOLA~\cite{foerster2017learning} can be leveraged to improve performance in such cases. However, when malfunctions or failures of the agents are not predictable, the challenge lies in enhancing the agents' adaptation capability. One approach to address this challenge is to guide the agents on how to cooperate under the current environmental circumstances, enabling them to make faster policy changes.

In this study, we propose a novel framework to enhance collaborative adaptation by steering the agents' behavior~\cite{findik2023influence} in scenarios where unexpected agent malfunction(s) occur. Our framework focuses on considering the inter-agent relationships, represented as a relational network, which captures the importance agents place on each other. By leveraging this relational network, agents can quickly adapt their learned behaviors to overcome unpredictable failures of their teammates. We specifically explore this concept using the VDN approach, a fast and powerful CTDE method for learning cooperative behaviors. Yet, it is crucial to emphasize that our framework of utilizing relationships to address unforeseen malfunctions can also be extended to other CTDE methods.


\section{Background}

\subsection{Markov Decision Process}
We characterized Decentralized Markov Decision Process as a tuple $\langle  \mathcal{S}, \mathcal{A}, \mathcal{R}, \mathcal{T}, \gamma\rangle$ where $s \in \mathcal{S}$ indicates the true state of the environment, the joint set of individual actions and rewards are represented by $\mathcal{A} \coloneqq \{a_1, a_2, \dots, a_n \}$, $\mathcal{R} \coloneqq \{r_1, r_2, \dots, r_n \}$, respectively,  $\mathcal{T} (s, A, s') \colon \mathcal{S} \times \mathcal{A}\times \mathcal{S} \mapsto [1,0]$ is the dynamics function defining the transition probability,  
$n$ is the the number of agents, and $\gamma\in[0,1)$ is the discount factor. 

\subsection{Value Function Factorization}

Value function factorization methods, which our proposed method build upon, adhere to the CTDE paradigm. These methods successfully tackle the non-stationarity issue in decentralized learning by employing centralized training and effectively address the scalability problem in centralized learning by adopting decentralized execution. Notably, QMIX~\cite{rashid2020monotonic} and VDN~\cite{sunehag2017value} serve as exemplary approaches in factorizing value functions.

QMIX and VDN both maintain a separate action-value, which defined as $Q_i(s, a_i)=\mathbb{E}[G | \mathcal{S}=s, \mathcal{A}=a_i]$ where $G$ denotes the return, for each agent $i \in \{ 1,...,n\}$. They merge these individual $Q_i$ values to obtain the central action value $Q_{\textrm{total}}$ using monotonicity and additivity. Specifically, VDN sums $Q_i$s to obtain $Q_{\textrm{total}}$, as 
$$Q_{\textrm{total}} = \sum_{i=1}^{n} Q_i(s, a_i),$$
while QMIX combines them using a state-dependent continuous monotonic function, as follows:
$$ Q_{\textrm{total}} = f_s(Q_1(s, a_1), ..., Q_n(s, a_n)), $$
where $\frac{\partial f_s}{\partial Q_i} \ge
0,  \forall i \in \{1, ..., n\}$.

These value function factorization methods rely on Deep Q-Network (DQN)~\cite{mnih2015human} to approximate the action-value function~$\hat{Q}_i(s, a_i, \theta_i)$ where $\theta_i$ is the weight vector. DQN is advantageous compared to tabular Q-learning as it can effectively handle high-dimensional state and action spaces by utilizing deep learning techniques. However, training DQN presents significant challenges due to instability and divergence resulting from updating the Q-network parameters in each step, violating the assumption of independently and identically distributed (i.i.d) data points. To tackle these challenges, Mnih et al.~\cite{mnih2015human} introduced techniques such as experience replay and fixed Q-target networks, which have now become standard in various deep reinforcement learning algorithms.

In brief, these value function factorization methods commonly utilize two deep Q-neural networks for each Q-function (i.e., each agent), namely the Prediction Neural Network (P-NN), and the fixed Target Neural Network (T-NN) which is essentially a copy of the P-NN from a previous iteration. Additionally, a replay memory is employed to store a large number of transitions experienced by the agent during its interactions with the environment. Each transition consists of a tuple $\langle s, a, r, s'\rangle$. To train the P-NN, a batch of transitions of size $b$ is sampled from memory, and the Temporal Difference (TD) error is calculated between the~$\hat{Q}_{\textrm{total}}^{\textrm{target}}$ and~$\hat{Q}_{\textrm{total}}^{\textrm{prediction}}$, as follows:
\begin{align}
\label{td_error}
e_{\textrm{TD}} = \sum_{i=1}^{b} [r_{\textrm{team}} + \gamma \max_{u'}(\hat{Q}_{\textrm{total}}(s', u', \theta_{t})) - \hat{Q}_{\textrm{total}}(s, u, \theta_{p})], 
\end{align}
where $r_{\textrm{team}}$ defined as the sum of rewards obtained by the agents, each having equal weights, $u$ denoted as the joint action of the agents, $\theta_p$ represents the weights of the P-NN and $\theta_t$
indicates the weights of the T-NN, which are regularly updated with $\theta_p$. And, $\theta_p$ are updated using an optimizer to minimize the $e_{\textrm{TD}}$. This process facilitates the coordination of agent actions towards maximizing the team reward. As a result, the key aspect of the CTDE paradigm becomes evident: the agent networks are trained using a centralized $Q_{\textrm{total}}$, while each agent's actions are determined by its own neural network, resulting in decentralized execution.

\section{Proposed Method}

In cooperative MARL, different team structures often result in multiple solutions of varying optimality. Value factorization methods and similar approaches aim to maximize team rewards and converge towards one of several solutions, potentially achieving the global optimum. The stochastic nature of agents' exploration can influence convergence towards a specific team behavior, particularly when multiple cooperation strategies exist with the same maximum total reward. However, in real-world scenarios, individual robots may encounter unexpected malfunctions (e.g., battery failure, rotation failure, etc.) after their policies have converged to a particular cooperative strategy, posing challenges for learning and adapting to new strategies without a deep understanding of the team structure.


To overcome these challenges, it would be beneficial to have a mechanism that considers inter-agent relationships and prioritizes assisting malfunctioning agents. This mechanism could improve team performance or accelerate adaptation by guiding agents' behavior towards either helping the malfunctioning agent solve its task or completing the task on its behalf. Unfortunately, the current cooperative MARL algorithms lack such a mechanism, making it more difficult and time-consuming to adapt to unforeseen malfunctions. To address this issue, we propose a novel framework called Collaborative Adaptation (CA). The CA framework enables agents to comprehend inter-agent relationships and select a cooperative strategy accordingly, allowing them to handle the adaptation of new environmental settings collaboratively. In our research, we explore and study this framework using the VDN algorithm, referred to as CA-VDN, due to its simplicity and effectiveness as a cooperative behavior learning approach.


The proposed framework employs a relational network in the form of a directed graph $\mathcal{G}=(\mathcal{V}, \mathcal{E}, \mathcal{W})$ to represent the relationships between agents. In this graph, each agent $i \in \{ 1,...,n\}$ is represented as a vertex $v_i$, $\mathcal{E}$ denotes the set of directed edges $e_{ij}$ directed from $v_i$ to $v_j$, and the weights of these edges are captured in the matrix $\mathcal{W}$, with elements $w_{ij}\in[0, 1]$ assigned to each edge. The direction and weight of the edges in the graph signify the importance or vested interest that agent $i$ places on the outcomes for agent $j$. Moreover, the framework modifies to MDP as $\langle \mathcal{S}, \mathcal{A}, \mathcal{R}, \mathcal{T}, \mathcal{G}, \gamma\rangle$ to incorporate $\mathcal{G}$. And, the $r_\textrm{team}$, used in~\eqref{td_error}, is calculated based on the relational network, as follows:
\begin{align}
\label{reward}
r_{\textrm{team}} = \sum_{i\in\mathcal{V}}^{} \sum_{j\in\mathcal{E}_i}^{} w_{ij}r_j,
\end{align}
where $\mathcal{E}_i$ denotes the set of vertex indices that have an edge directed from $v_i$, and $r_j$ is the reward of the agent represented by $v_j$. This allows for the agents to follow a cooperative strategy that assists the malfunction agent since they place extra importance on its reward. The pseudo-code for the CA framework can be found in Algorithm~\ref{alg:CA}.



\renewcommand{\algorithmicrequire}{\textbf{Input:}}
\renewcommand{\algorithmicensure}{\textbf{Output:}}

\begin{algorithm}[t]
    \caption{Collaborative Adaptation}
    \label{alg:CA}
    \SetAlgoLined
    \SetKwInOut{Input}{input}
    \SetKwInOut{Output}{output}
    \DontPrintSemicolon
    
    \Input{P-NN, $\hat{Q}^{\textrm{prediction}}$; T-NN, $\hat{Q}^{\textrm{target}}$; relational network, $G$; batch size, $b$; number of iterations for updates, $m$; update frequency of T-NN, $k$}
    \ForEach{episode}{
        Initialize $s$\;
        \ForEach{step of episode}{
            Choose $a$ from $s$ using policy derived from $\hat{Q}^{\textrm{prediction}}$ (with $\varepsilon$-greedy)\;
            Take action $a$, observe $r$, $s'$\;
            Store $s$, $a$, $r$, $s'$ in memory\;
            $s \leftarrow s'$\;
        }
        \For{$i=1, \ldots, m$}{
            $S$, $A$, $R$, $S'$ $\leftarrow$ sample chunk, size of $b$, from memory\;
            $Q^\textrm{prediction}_\textrm{values} \leftarrow \hat{Q}^{\textrm{prediction}}$($S$) \;
            $Q^\textrm{prediction}_\textrm{values} \leftarrow$ action $A$ of $Q^\textrm{prediction}_\textrm{values}$ of every agent in every sample\;
            $Q^{\textrm{prediction}}_{\textrm{total}} \leftarrow$ sum $Q^{\textrm{prediction}}$ per sample\;
            $Q^\textrm{target}_\textrm{values} \leftarrow \hat{Q}^{\textrm{target}}$($S'$)\;
            $Q^\textrm{target} \leftarrow$ max of $Q^\textrm{target}_\textrm{values}$ of every agent in every sample \;
            $Q^{\textrm{target}}_{\textrm{total}} \leftarrow$ sum $Q^{\textrm{target}}$ per sample\;
            $R^{\textrm{team}} \leftarrow$ use~\eqref{reward} with $G$ and $R$\;

            $loss \leftarrow$ use~\eqref{td_error} with $R^{\textrm{team}}$, $Q^{\textrm{target}}_{\textrm{total}}$, $Q^{\textrm{prediction}}_{\textrm{total}}$\;
        
            Backpropagate the loss to the parameters of $\hat{Q}^{\textrm{prediction}}$\;
        }
        Update the parameters of $\hat{Q}^{\textrm{target}}$ with the parameters of $\hat{Q}^{\textrm{prediction}}$ foreach $k^{\textrm{th}}$ episode\;

    }

\end{algorithm}

To identify malfunctioning agents and determine when these malfunctions occur, and facilitate changes in inter-agent relations to support these agents, a mechanism -- malfunction trigger -- has been implemented to track individual agents' rewards. The underlying assumption is that agents may experience malfunctions after converging on a specific behavior, highlighting the challenges faced by existing cooperative MARL algorithms in altering already converged behaviors in response to unpredictable failures. When the malfunction trigger observes a significant decrease in an individual agent's reward over a certain number of episodes, it signals this information to the framework, indicating the presence of malfunctioning agent(s). Upon receiving this information, the framework dynamically adjusts the relational network, leading other agents to assign importance to the malfunctioning agent(s) by modifying the weights of the corresponding edges. Additionally, the framework resets the exploration process to allow the agent to discover new cooperative strategies based on the updated relationships. It's important to note that during the comparison of results with other methods, the exploration parameter of these methods is also reset to enable them to explore anew.


\begin{figure}[t]
\centering
  \includegraphics[width=\linewidth]{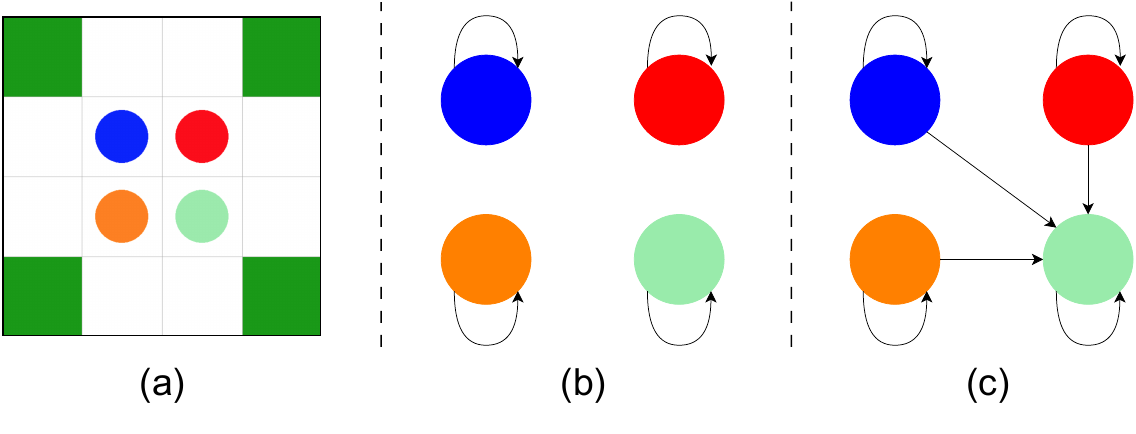}
  \caption{\small{(a) multi-agent grid-world environment with four agents. (b-c) Relational networks employed in CA-VDN}} 
  \label{fig:relations}
\end{figure}

\section{Experiments}

\begin{figure}[b]
\centering
\includegraphics[width=\linewidth]{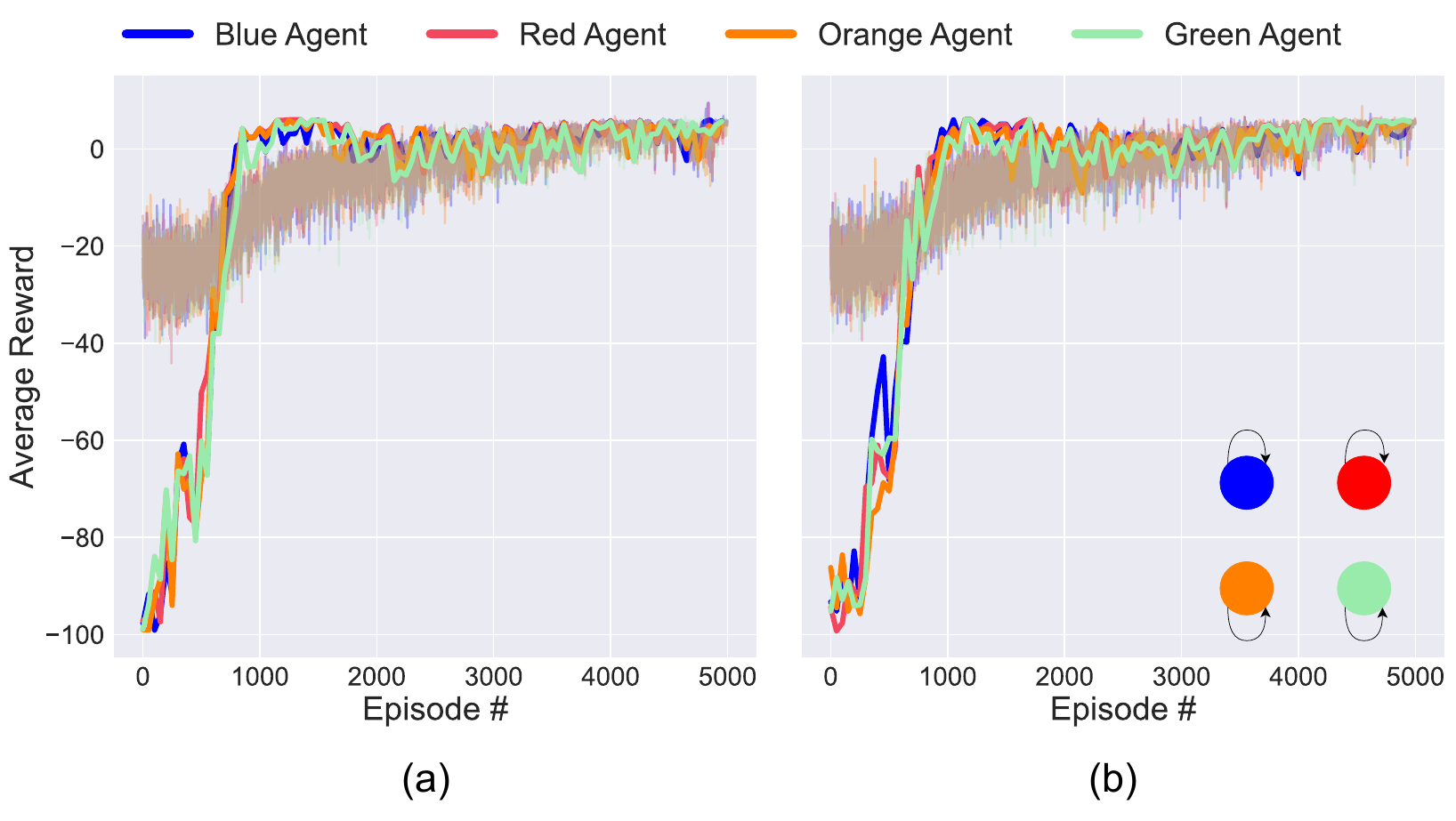}
  \caption{\small{Results before malfunction. (a) VDN, (b) CA-VDN with relational network in Fig.~\ref{fig:relations}(b).}} 
  \label{fig:first_part}
\end{figure}

\subsection{Environment}

To evaluate the effectiveness of the proposed approach in influencing agents' behaviors and enhancing their adaptation to unforeseen failures of the agent(s), we conducted experiments using the CA-VDN and VDN algorithms in a multi-agent grid-world environment. The environment is represented as a 4x4 grid with four agents and four undedicated resources, as illustrated in Fig. (a).

In this environment, the objective of each episode is for the agents to consume all the resources by visiting their respective locations. To achieve this, the agents have five possible actions: move up, down, left, right, or stay idle. Additionally, they can engage in a special action called \textit{push}, which allows them to push adjacent agents, provided that the pushing agent takes a non-idle action towards the pushed agent, who must be idle. As a result of a \textit{push}, the pushing agent remains in place while the other agent moves one space in the direction of the push.

Upon successfully consuming a resource, the consumer agent receive a reward of $+10$ and each resource can only be consumed once. Nevertheless, the agent incurs an individual penalty of $-1$ for every time-step per unconsumed resource, except when they are occupying a resource location, which serves as a safe spot. The episode terminates either when all the resources are consumed or when the maximum time steps are reached.



We intentionally designed this environment to be solvable by VDN while also highlighting the challenges that unexpected malfunctions can bring, even in a seemingly simple setting. Furthermore, our goal is to showcase how the integration of relationships between agents into the learning process can effectively overcome these challenges.


\subsection{Models and Hyperparameters}

In our experimental setup, we utilized a Multi-Layer Perceptron (MLP) with two hidden layers, each containing $128$ neurons, and using the ReLU activation function. To train each agent's prediction model, we conducted $m=10$ iterations per episode, using batches of size $b=32$ randomly sampled from a replay memory with a capacity of $50$k time-steps. The optimization was performed using the \textit{Adam} optimizer with a learning rate of $0.001$, and the squared TD-error served as the loss function.

To maintain stability during training, we updated the weights of the target network with the prediction network's weights every $k=200$ episodes. For exploration, we employed the $\varepsilon$-greedy method, with $\varepsilon$ linearly decreasing over time. Lastly, we set the discount factor ($\gamma$) to 0.99 to account for future rewards in the reinforcement learning process.

\begin{table}[b]
\centering
\caption{
Average reward with 95\% confidence intervals for ten runs after training completed.
}
\resizebox{\linewidth}{!}{
\begin{tabular}{ccc|cc}
\hline
             & \multicolumn{2}{c|}{Before Malfunction} & \multicolumn{2}{c}{After Malfunction} \\ \cline{2-5} 
             & \multicolumn{1}{c|}{VDN}  & CA-VDN     & \multicolumn{1}{c|}{VDN} & CA-VDN    \\ \cline{2-5} 
Blue Agent   & 5.80±0.25                 & 5.50±0.64  & -74.20±18.14                & \textbf{6.90±0.19} \\
Red Agent    & 5.50±0.50                 & 5.70±0.40  & -63.60±20.26                & \textbf{9.90±0.19} \\
Orange Agent & 5.20±0.67                 & 5.40±0.79  & -66.50±21.47                & \textbf{9.70±0.40} \\
Green Agent  & 5.70±0.40                 & 5.60±0.74  & -35.70±28.60                & \textbf{10.50±0.93} \\ \hline
\end{tabular}
}
\label{experiment_results}
\end{table}

\subsection{Results \& Discussion}

\begin{figure}[t]
\centering
\includegraphics[width=\linewidth]{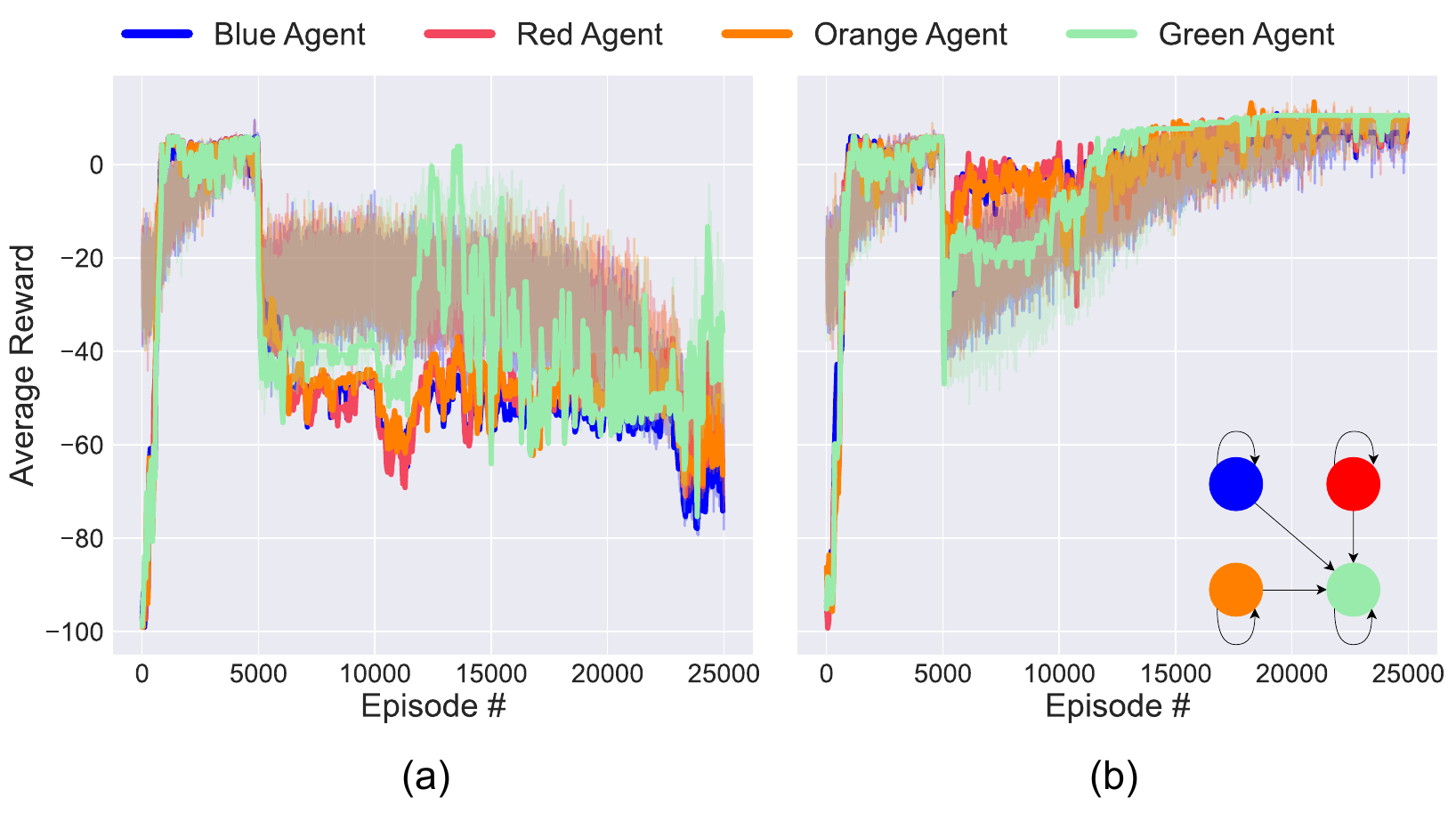}
  \caption{\small{Results with malfunction. (a) VDN, (b) CA-VDN with relational network in Fig.~\ref{fig:relations}(c).}} 
  \label{fig:second_part}
\end{figure}

The experimental results, presented in Fig.~\ref{fig:first_part} and Fig.~\ref{fig:second_part}, show the average training reward over $10$ runs, represented by the shaded regions, as well as the average test rewards of the agents indicated by the solid lines. The test rewards are evaluated based on a greedy strategy, interrupting the training process every $50$ episodes to assess individual agent rewards. During each run, at the five thousandth episode, we simulate a malfunction that prevents the green agent (refer to Fig.~\ref{fig:relations}(a)) from moving.
It is crucial to highlight that this malfunction was not anticipated by the algorithms.


During the initial phase, when all agents are fully functional, both VDN and CA-VDN, which incorporate the relational network illustrated in Fig.~\ref{fig:relations}(b), demonstrate comparable performance. This similarity is evident in the agents' average rewards, as depicted in Fig.~\ref{fig:first_part}(a) and Fig.~\ref{fig:first_part}(b) respectively, and they eventually converge to the same behavior. The resemblance in performance arises from the fact that in VDN, each agent contributes their reward equally to the team's overall reward, aligning with the utilization of the self-interested relational network. Essentially, the agents share resources among themselves based on their proximity to those resources and subsequently consume them. The similarity in individual rewards for both algorithms after the initial phase concludes is demonstrated in Table~\ref{fig:first_part}.


At the five thousandth episode, our framework's malfunction trigger detects the occurrence of a malfunction. As VDN lacks a support mechanism that considers inter-agent relationships, the only available option is to reset the $\varepsilon$ value to increase exploration. Despite this attempt, as shown in Fig.~\ref{fig:second_part}(a), VDN still faces challenges in recovering from malfunction scenarios.

On the other hand, during the reset of $\varepsilon$, we also modify the applied relational network from Fig.~\ref{fig:relations}(b) to Fig.~\ref{fig:relations}(c). This alteration ensures that other agents place importance on the malfunctioning agent. As a result, agents can adapt faster to the new condition, as indicated in Fig.~\ref{fig:second_part}(b). To evaluate the effectiveness, we present the numeric results of each agent's individual reward in Table~\ref{experiment_results} for both before and after the malfunction.

Overall, it is essential to highlight that agents trained with CA-VDN can learn together to recover from unforeseen malfunctions, a capability that VDN lacks even after 20k episodes following the occurrence of a malfunction.


\section{Conclusion and Future Work}

We propose a novel framework that incorporates inter-agent relationships into agents' learning, enabling agents to recover from unforeseen malfunctions as a team. Our experiments validated the effectiveness of our approach in faster adaptation to the environment in the face of unexpected robot failures. As a next step, we aim to conduct additional experiments in more complex environments that involve multiple agents with different malfunctions and compare the performance of our algorithm with other state-of-the-art methods.

\section*{Acknowledgement}
This work is supported in part by NSF (IIS-2112633) and
the Army Research Lab (W911NF20-2-0089).

\addtolength{\textheight}{-20.2cm}

\bibliographystyle{IEEEtran}
\bibliography{IEEEabrv, refs}

\end{document}